\documentclass[11pt]{article}
\usepackage[a4paper]{geometry}
\usepackage{clic2017}
\usepackage{times}
\usepackage{url}
\usepackage{latexsym}
\usepackage{booktabs,multirow}
\usepackage{todonotes}
\usepackage{lipsum}
\usepackage[utf8]{inputenc}

\title{Datasets and Models for Authorship Attribution \\ on Italian  Personal Writings}

\newcommand\blfootnote[1]{%
  \begingroup
  \renewcommand\thefootnote{}\footnote{#1}%
  \addtocounter{footnote}{-1}%
  \endgroup
}

\author{\bf {Gaetana Ruggiero}$^{\bullet}$$^{\diamond}$, \bf{Albert Gatt$^{\bullet}$, Malvina Nissim$^{\diamond}$}\\ 
~
\\
  \textsuperscript{$\bullet$}Institute of Linguistics and Language Technology, University of Malta, Malta \\ \textsuperscript{$\diamond$}Center for Language and Cognition, University of Groningen, The Netherlands\\
\normalsize{\texttt{garuggiero@gmail.com}}, \normalsize{\texttt{albert.gatt@um.edu.mt}},
   \normalsize{\texttt{m.nissim@rug.nl}}}

\date{}

\begin{document}
\maketitle
\begin{abstract}
 Existing research on Authorship Attribution (AA) focuses
  on  texts for which a lot of data is available (e.g novels), mainly in English.
  We approach AA via Authorship Verification
  on short  Italian texts in
  two novel datasets, and 
  analyze the interaction between genre, topic, gender and length. Results show that AV is feasible even with little data, but  more  evidence  helps.   Gender  and  topic  can be indicative clues, and if not controlled for, they might overtake more specific aspects of personal style.
 
\end{abstract}

\section{Introduction and Background}\label{introduction}

Authorship Attribution (AA) is the task of identifying authors by their writing style. In addition to being a tool for studying individual language choices, AA is useful for many real-life applications, such as plagiarism detection  \cite{stamatatos2011plagiarism},  multiple accounts detection \cite{tsikerdekis2014multipleaccount}, and online security \cite{yang2014onlinesecurity}. 

Most work on AA focuses on English, on relatively long texts such as novels and articles 
\cite{juola2015rowling} where personal style could be mitigated due to editorial interventions.
Furthermore, in many real-world applications the texts of disputed authorship tend to be short \cite{omar2019shortness}.

\blfootnote{Copyright ©2020 for this paper by its authors. Use permitted under Creative Commons License Attribution 4.0 International (CC BY 4.0).}
The PAN 2020 shared task was originally meant to investigate multilingual AV in fanfiction, focusing on Italian, Spanish, Dutch and English \cite{pan2020}. However, the datasets were eventually restricted to  English only, to maximize the amount of available training data \cite{overviewpan2020}, emphasizing the difficulty in compiling large enough datasets for less-resourced languages.

AA research in Italian has largely focused on the single case of Elena Ferrante \cite{tuzzi2018ferrante} \footnote{\footnotesize{https://www.newyorker.com/culture/cultural-comment/the-unmasking-of-elena-ferrante}}. The present work seeks a more realistic take, using more diverse, user-generated data namely web forums comments and diary fragments, thereby introducing two novel datasets for this task: \textit{ForumFree} and \textit{Diaries}.

We cast the AA problem as \textit{authorship verification} (AV). Rather than identifying the specific author of a text (the most common task in AA), AV aims at determining whether two texts were written by the same author or not \cite{koppel2004oneclass,koppel2009computational}.

The GLAD system of \newcite{glad} was specifically developed to solve AV problems, and has been shown to be highly adaptable to new datasets \cite{halvani2018unary}. GLAD uses an SVM with a variety of features including character level ones, which have proved to be most effective for AA tasks \cite{stamatatos2009survey,moreau2015author,glad}, and is freely available. Moreover, \newcite{pan2019} show that many of the best models for authorship attribution are based on Support Vector Machines. Hence we adopt GLAD in the present study.

More specifically, we run GLAD on our datasets and study the interaction of four different dimensions: topic, gender, amount of evidence per author, and genre. In practice, we design intra-topic, cross-topic, and cross-genre experiments, controlling for gender and amount of evidence per author.
The focus on cross-topic and cross-genre AV is in line with the PAN 2015 shared task \cite{pan2015overview}; this setting has been shown to be more challenging than the task definitions of previous editions \cite{pan2013overview,pan2014overview}.

\paragraph{Contributions}
We advance AA for Italian introducing two novel datasets, \textit{ForumFree} and \textit{Diaries}, which contribute to enhance the amount of available Italian data suitable for AA tasks.\footnote{Further information about the datasets can be found at https://github.com/garuggiero/Italian-Datasets-for-AV } 

Running a battery of experiments on personal writings, we show that AV is feasible even with little data, but more evidence helps. Gender and topic can be indicative clues, and if not controlled for, they might overtake more specific aspects of personal style. 

\section{Data}
\label{datasets}
For the present study, we introduce two novel datasets, \textit{ForumFree} and \textit{Diaries}.  Although already compiled \cite{maslennikova2019quanti}, the original ForumFree dataset was not meant for AA. Therefore, we reformat it following the PAN format\footnote{https://pan.webis.de/clef15/pan15-web/authorship-verification.html}. The dataset contains web forum comments taken from the ForumFree platform\footnote{https://www.forumfree.it/}, and the subset used in this work covers two topics, \textit{Medicina Estetica} (``Aesthethic Medicine'') and \textit{Programmi Tv} (``Tv Programmes''; \textit{Celebrities} in the original dataset). A third subset, \textit{Mix}, is the union of the first two. The Diaries dataset is originally assembled for the present study, and contains a collection of diary fragments included in the project \textit{Italiani all'estero: i diari raccontano} (``Italians abroad:  the diaries narrate'').\footnote{https://www.idiariraccontano.org} For  Diaries, no topic classification has been taken into account.
 Table~\ref{tab:datasets_overview} shows an overview of the datasets.

\begin{table}[h] 
\small
\centering
\resizebox{\columnwidth}{!}{
\begin{tabular}{l|rrrrrrr}
\toprule
\multicolumn{1}{c|}{\textbf{Subset}} & \multicolumn{3}{c}{\textbf{\# Authors}} & \multicolumn{1}{c}{\textbf{\# Docs}} & \multicolumn{1}{c}{\textbf{\begin{tabular}[c]{@{}c@{}}W/A \end{tabular}}} & \multicolumn{1}{c}{\textbf{\begin{tabular}[c]{@{}c@{}}D/A \end{tabular}}} & \multicolumn{1}{c}{\textbf{\begin{tabular}[c]{@{}c@{}}W/D\end{tabular}}} \\
 & \multicolumn{1}{l}{\textbf{F}} & \multicolumn{1}{l}{\textbf{M}} & \multicolumn{1}{l}{\textbf{Tot}} & \multicolumn{1}{l}{} & \multicolumn{1}{l}{} & \multicolumn{1}{l}{} & \multicolumn{1}{l}{} \\ \midrule
Med Est & 33 & 44 & 77 & 56198 & 63 & 661 & 48 \\ \midrule
Prog TV & 78 & 71 & 149 & 153019 & 32 & 812 & 22 \\ \midrule
Mix & 111 & 115 & 276 & 209217 & 41 & 791 & 29 \\ \midrule
Diaries & 77 & 188 & 275 & 1422 & 462 & 5 & 477 \\
\bottomrule
\end{tabular}
}
 \caption{Overview of the datasets. W/A = Avg words per author; D/A = Avg docs per author; W/D = Avg words per doc.}

\label{tab:datasets_overview}
\end{table}

\subsection{Preprocessing}\label{preprocessing}

For the ForumFree dataset, comments which only contained the word \textit{up}, commonly used on the internet to give new visibility to a post that was written in the past, were removed from the dataset, together with their authors when this was the only text associated with them. 

 The stories narrated in the diaries are of a very personal nature, which means that many proper nouns and names of locations are used. To avoid relying on these explicit clues, which are strong but not indicative of personal writing style, we perform Named Entity Recognition (NER), using \textit{spaCy} \cite{honnibal2015spacy}. Person names, locations and organizations were replaced by their corresponding labels, namely \textit{PER}, \textit{LOC}, \textit{ORG}. The fourth label used by  \textit{spaCy}, MISC (miscellany), was not considered; dates were also not normalized. Moreover, 
a separate set of experiments was performed by \textit{bleaching} the diary texts prior to their input to the GLAD system. The bleaching method was proposed by \newcite{van2018bleaching} in the context of cross-lingual Gender Prediction, and consists of transforming tokens into an abstract representation that masks lexical forms while maintaining key features. We only use 4 of the 6 original features. \textit{Shape} transforms uppercase letters  into `U', lowercase ones into `L', digits into `D', and the rest into `X'. \textit{PunctA} replaces emojis with `J', emoticons with `E', punctuation with `P' and one or more alphanumeric characters with a single `W'. \textit{Length} represents a word by the number of its characters.
\textit{Frequency} corresponds to the \textit{log} frequency of a token in the dataset. The features are then concatenated. The word `House' would be rewritten as `ULLLL W 05 6'.

\begin{figure*}[t]
    \centering
    \includegraphics[width=.8\textwidth]{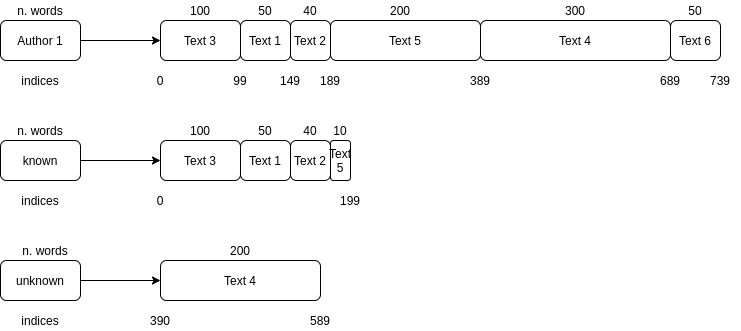}
    \caption{Example of the creation of known and unknown documents for the same author when considering 400 words per author.}
    \label{fig:ku_pair_creation}
\end{figure*}

\subsection{Reformatting}\label{reformatting}
We reformat both datasets in order to make them suitable for AV. The data is divided into so-called \textit{problems}:  each problem is made of a known and an unknown text of equal length. 

To account for the shortness of the texts and to avoid topic biases that would derive by taking consecutive text as known and unknown fragments, all the documents written by the same author are first shuffled and then concatenated into a single string. The string is split into two spans containing the same number of words, so that the words contained in the unknown span come from subsets of texts which are different from the ones that form the known one. An example of this process is displayed in Figure~\ref{fig:ku_pair_creation}. 
Rather than being represented by individual productions, 
each author is therefore represented by a \textit{set} of texts, whose original sequential order has been altered. Each known text is paired with an unknown text from the same author. To create negative instances, given a dataset with multiple problems, one can (i) make use of external documents (\textit{extrinsic} approach \cite{seidman2013authorship,koppel2014determining}), or 
(ii) use fragments collated from all authors in the training data, except the target author (\textit{intrinsic} approach). We create negative instances with an intrinsic approach. More specifically, following \newcite{masterthesisdwyer}, the second half of the unknown array is shifted by one, so that the texts of the second half of the known array are paired with a \textit{different-author} text in the unknown array. In this way, the label distribution is balanced.

\section{Method}\label{sec:method}

Given a pair of known and unknown fragments (KU pair), the task is to predict whether they are written by the same author or not. In designing our experiments, we control for topic, gender, amount of evidence, and genre. The latter is fostered by the diverse nature of our datasets.

\paragraph{Topic} Maintaining the topic roughly constant should allow stylistic features to gain more discriminative value. We design intra-topic (IT) and cross-topic experiments (CT). In IT, we distinguish same- and different-topic KU pairs. In same-topic, we train and test the system on KU pairs from the same topic. In different-topic, we include the Mix set and the diaries. Since we train and test on a mixture of topics and there can be topic overlap, these are not truly cross-topic, and we do not consider them as such.

Given that no topic classification is available for the diaries, the CT experiments are
only performed on the ForumFree dataset. We train the system on Medicina Estetica and test it on Programmi Tv, and vice versa. 

\paragraph{Gender} Previous work has shown that similarity can be observed in writings of people of the same gender \cite{profilingpaper,pan2017}.\footnote{Binary gender is a simplification of a much more nuanced situation in reality. Following previous work, we adopt it for convenience.} In order to assess the influence of same vs different gender in AA, we consider three gender settings: only female authors and only male authors (\textit{single-gender}), and \textit{mixed-gender}, where the known and unknown document can be either written by two authors of the same gender, or by a male and a female author. In dividing the subsets according to the gender of the authors, we consider gender implicitly. However, we also perform experiments adding gender as feature to the instance vectors, indicating both the gender of the known and unknown documents' authors and whether or not the gender of the authors is the same. 

\paragraph{Evidence}
Following \newcite{feiguina2007smalltexts}, we experiment with KU pairs of different sizes, i.e. with 400, 1\,000, 2\,000 and 3\,000 words per author. Each element of the KU pair is thus made up of 200, 500, 1\,000 and 1\,500 words respectively. To observe the effect of the different text sizes on the classification, we manipulate the number of instances in training and test, so that the same authors are included in all the different word settings of a single topic-gender experiment.

\paragraph{Genre} We perform cross-genre experiments (CG) by training on ForumFree and testing on the Diaries, and vice versa.

\paragraph{Splits and Evaluation}  We train on 70\% and test on 30\% of the instances. However, since we are controlling for gender and topic, the number of instances contained in the training and test sets varies in each experiment.
We keep the test sets stable across IT, CT and CG experiments, so that we can compare results. Following the PAN evaluation settings  \cite{pan2015overview}, we use three metrics. $c@1$  takes into account the number of problems left unanswered and rewards the system when it classifies a problem as unanswered rather than misclassifying it. Probability scores are converted to binary answers: every score greater than 0.5 becomes a positive answer, every score smaller than 0.5 corresponds to a negative answer and every score which is exactly 0.5 is considered as an unanswered problem. The $AUC$ measure corresponds to the area under the ROC curve \cite{roccurve}, and tests the ability of the system 
to rank scores properly, assigning low values to negative problems and high values to positive ones \cite{pan2015overview}. The third measure is 
the product of $c@1$ and $AUC$.

\paragraph{Model}
We run all experiments using GLAD \cite{glad}. This is an SVM with \textit{rbf} kernel, implemented using Python's \textit{scikit-learn} \cite{pedregosa2011scikit} library and 
NLTK \cite{bird2009nltk}. GLAD was designed to work with 24 different features, which take into account stylometry, entropy and data compression measures. We compare GLAD to a simple baseline which randomly assigns a label from the set of possible labels (i.e. `YES' or `NO') to each test instance. 

Our choice fell on GLAD for a variety of reasons. As a general observation, even in later challenges, SVMs have proven to be the most effective for AA tasks \cite{pan2019}. More specifically, in a survey of freely available AA systems, GLAD showed best performance and especially high adaptability to new datasets \cite{halvani2018unary}. Lastly, \newcite{masterthesisdeVries} has explored fine-tuning a pre-trained model for AV in Dutch, a less-resourced language compared to English. He found that fine-tuning BERTje (a Dutch monolingual BERT-model, \cite{de2019bertje}) with PAN~2015 AV data \cite{pan2015overview},  failed to outperform a majority baseline \cite{masterthesisdeVries}. He concluded that Tranformer-encoder models might not suitable for AA tasks, since they will likely overfit if the documents contain no reliable clues of authorship \cite{masterthesisdeVries}. 

\section{Results and Discussion}
The number of experiments is high due to the interaction of the dimensions we consider. 

Tables~\ref{tab:it_mix_mixed} and \ref{it_diaries_mixed} only include the mixed-gender results of the IT experiments on Mix (which corresponds to the entire ForumFree dataset used for this study) and Diaries, respectively. Results concerning all dimensions considered are anyway discussed in the text. We refer to the combined score. Since the baseline results are different for each setting, we do not include them. However, all models perform consistently above their corresponding baseline. 

For the Mix topic, we achieved 0.966 with 96 authors in total and 3\,000 words (Table~\ref{tab:it_mix_mixed}). For the diaries, we achieved 0.821 with 46 authors in total and 3\,000 words each (Table~\ref{it_diaries_mixed}).\footnote{Using a bleached representation of the texts, the score increased by 0.36} 
Although the training and test sets are of different sizes for both datasets, more evidence seems to help the model to solve the problem. 

In the IT experiments, the highest score for Medicina Estetica is 0.923, with 41 authors in total and 1\,000 words per author, and for Programmi Tv 0.944, with 59 authors and 3\,000 words each. In the CT setting, the scores stay basically the same in both directions.
In CG, when training on the diaries and testing on Mix, we obtain the same score when training on Mix with 3\,000 words. When training on Mix and testing on Diaries,  we achieved 0.737 on the same test set, and 0.748 with 1\,000 words per instance.

\begin{table*}[t]
\centering
\begin{tabular}{r|r|rr|rrrrrr}

\toprule
\multirow{2}{*}{\textbf{\# W/A}} &

\multirow{2}{*}{\textbf{\# Auth}} & \multicolumn{2}{c|}{\textbf{\# Problems}} &  \multicolumn{6}{c}{\textbf{\begin{tabular}[c]{@{}c@{}}Eval\end{tabular}}} \\ \cline{3-10}
\multicolumn{1}{c|}{\textbf{}} &
\multicolumn{1}{c|}{\textbf{}} & \multicolumn{1}{c}{\textbf{Train}} & \multicolumn{1}{c|}{\textbf{Test}} &  \multicolumn{1}{c}{\textbf{C}} & \multicolumn{1}{c}{\textbf{I}} & \multicolumn{1}{c}{\textbf{U}} & \multicolumn{1}{c}{\textbf{c@1}} & \multicolumn{1}{c}{\textbf{AUC}} & \multicolumn{1}{c}{\textbf{*}} \\ \midrule
\textbf{\textbf{400}} & \multicolumn{1}{r|}{127} & 88 & \multicolumn{1}{r|}{39} & 33 & 6 & 0 & 0.846 & 0.947 & 0.801 \\
{\textbf{1 000}} & \multicolumn{1}{r|}{109} & 76 & \multicolumn{1}{r|}{33} & 30 & 3 & 0 & 0.909 & 0.926 & 0.842 \\
{\textbf{2 000}} & \multicolumn{1}{r|}{100} & 70 & \multicolumn{1}{r|}{30} & 29 & 1 & 0 & \textbf{0.967} & 0.995 & 0.962 \\
{\textbf{3 000}} & \multicolumn{1}{r|}{96} & 67 & \multicolumn{1}{r|}{29} & 28 & 1 & 0 & 0.966 & \textbf{1.000} & \textbf{0.966} \\ \bottomrule
\end{tabular}
\caption{Training and test set configurations and IT evaluation scores on Mix texts written by female and male authors. \textit{C,I} and \textit{U} are Correct, Incorrect, Unanswered problems.}
\label{tab:it_mix_mixed}
\end{table*}

\begin{table*}[ht]
\centering
\begin{tabular}{r|r|rr|rrrrrr}
\toprule

\multirow{2}{*}{\textbf{\# W/A}} &

\multirow{2}{*}{\textbf{\# Auth}} & \multicolumn{2}{c|}{\textbf{\# Problems}} &  \multicolumn{6}{c}{\textbf{\begin{tabular}[c]{@{}c@{}}Eval\end{tabular}}} \\ \cline{3-10}

\multicolumn{1}{c|}{\textbf{}} &
\multicolumn{1}{c|}{\textbf{}} & \multicolumn{1}{c}{\textbf{Train}} & \multicolumn{1}{c|}{\textbf{Test}} &  \multicolumn{1}{c}{\textbf{C}} & \multicolumn{1}{c}{\textbf{I}} & \multicolumn{1}{c}{\textbf{U}} & \multicolumn{1}{c}{\textbf{c@1}} & \multicolumn{1}{c}{\textbf{AUC}} & \multicolumn{1}{c}{\textbf{*}} \\ \midrule
\multicolumn{1}{r|}{\textbf{400}} & \multicolumn{1}{r|}{229} & 160 & \multicolumn{1}{r|}{69} & 47 & 21 & 1 & 0.691 & 0.725 & 0.500 \\
\multicolumn{1}{r|}{\textbf{1 000}}& \multicolumn{1}{r|}{180} & 126 & \multicolumn{1}{r|}{54} & 43 & 11 & 0 & 0.796 & 0.891 & 0.709 \\
\multicolumn{1}{r|}{\textbf{2 000}} & \multicolumn{1}{r|}{98} & 68 & \multicolumn{1}{r|}{30} & 25 & 5 & 0 & \textbf{0.833} & 0.905 & 0.754 \\
\multicolumn{1}{r|}{\textbf{3 000}} & \multicolumn{1}{r|}{46} & 32 & \multicolumn{1}{r|}{14} & 12 & 2 & 0 & 0.857 & \textbf{0.958} & \textbf{0.821}\\
\bottomrule
\end{tabular}
\caption{Training and test configurations and IT evaluation scores on diaries made of NE converted text written by both genders. \textit{C,I} and \textit{U} are Correct, Incorrect, Unanswered problems.}
\label{it_diaries_mixed}
\end{table*}

 \paragraph{Discussion}

When more variables interact in the same subset, as in  mixed-gender sets of the ForumFree and Diaries dataset, we found that the classifier uses the implicit gender information. Indeed, it achieves slightly better scores in mixed-gender settings than in female- and male-only ones, suggesting that the classifier might be using internal clustering of the data rather than writing style characteristics. This also explains why results are higher in Mix than in separate topics, because the classifier can use topic information. 

We also observe that by adding gender as an explicit feature in topic- and gender-controlled subsets, GLAD uses this information to improve classification, especially in mixed-gender scenarios.

Although previous research demonstrated that CT and CG experiments are harder than IT ones \cite{sapkota2014outoftopicdata,pan2015overview},  in our case the scores for the three settings are comparable. However, since we only performed CT and CG experiments on mixed-gender subsets, the gender-specific information might have also played a role in this process (see above).

Overall, the experiments show
that using a higher number of words per author is preferable. Although 3\,000 words seems to be optimal for most settings, in the large number of experiments that we carried out (not all included in this paper) we also observed that lower amounts of words also led to comparable results. This aspect will require further investigation.

\section{Conclusion}

We experimented with AV on Italian forum comments and diary fragments.
We compiled two datasets and performed experiments which
considered
the interaction among topic, gender, length and genre. Even when the texts are short and present more individual variation than traditional texts used in AA, AV is a feasible task, but having more evidence per author improves classification. While making the task more challenging, controlling for gender and topic ensures that the system prioritizes authorship over different data clusters. Although the datasets used are intended for
AV problems, they can be easily adapted to other AA tasks. 
We believe this to be one of the major contributions of our work, as it can help to advance the up-to-now limited AA research in Italian.

\section*{Acknowledgments}

The ForumFree dataset was a courtesy  of  the  Italian Institute of Computational Linguistics “Antonio Zampolli” (ILC) of Pisa.\footnote{http://www.ilc.cnr.it/}

\bibliographystyle{acl.bst}
\bibliography{acl2014}

\end{document}